\documentclass[runningheads]{llncs}
\usepackage{graphicx}
\usepackage{comment}
\usepackage{amsmath,amssymb} %
\usepackage{color}
\usepackage{cite}

\usepackage{enumitem}
\usepackage{multirow}
\DeclareMathAlphabet\mathbfcal{OMS}{cmsy}{b}{n}
\usepackage{mathtools}

\DeclarePairedDelimiterX{\infdivx}[2]{(}{)}{%
  #1\;\delimsize\|\;#2%
}

\DeclarePairedDelimiter{\norm}{\lVert}{\rVert}

\begin{document}
\pagestyle{headings}
\mainmatter
\def\ECCVSubNumber{2976}  %

\title{Preserving Semantic Neighborhoods for Robust Cross-modal Retrieval} %

\titlerunning{Preserving Semantic Neighborhoods for Robust Cross-modal Retrieval}
\author{Christopher Thomas\orcidID{0000-0002-3226-396X} \and
Adriana Kovashka\orcidID{0000-0003-1901-9660}}
\authorrunning{Thomas and Kovashka}
\institute{University of Pittsburgh, Pittsburgh PA 15260, USA\\ 
\email{\{chris, kovashka\}@cs.pitt.edu}}
\maketitle

\begin{abstract}
The abundance of multimodal data (e.g. social media posts) has inspired interest in cross-modal retrieval methods. 
Popular approaches rely on a variety of metric learning losses, which prescribe what the proximity of image and text should be, in the learned space. 
However, most prior methods have focused on the case where image and text convey redundant information; in contrast, real-world image-text pairs convey complementary information with little overlap.
Further, images in news articles and media portray topics in a visually diverse fashion; thus, we need to take special care to ensure a meaningful image representation. We propose novel within-modality losses which encourage semantic coherency in both the text and image subspaces, which does not necessarily align with visual coherency. Our method ensures that not only are paired images and texts close, but the expected image-image and text-text relationships are also observed. Our approach improves the results of cross-modal retrieval on four datasets compared to five baselines.
\end{abstract}

\section{Introduction}

\begin{figure*}[t]
    \centering
    \includegraphics[width=1\textwidth]{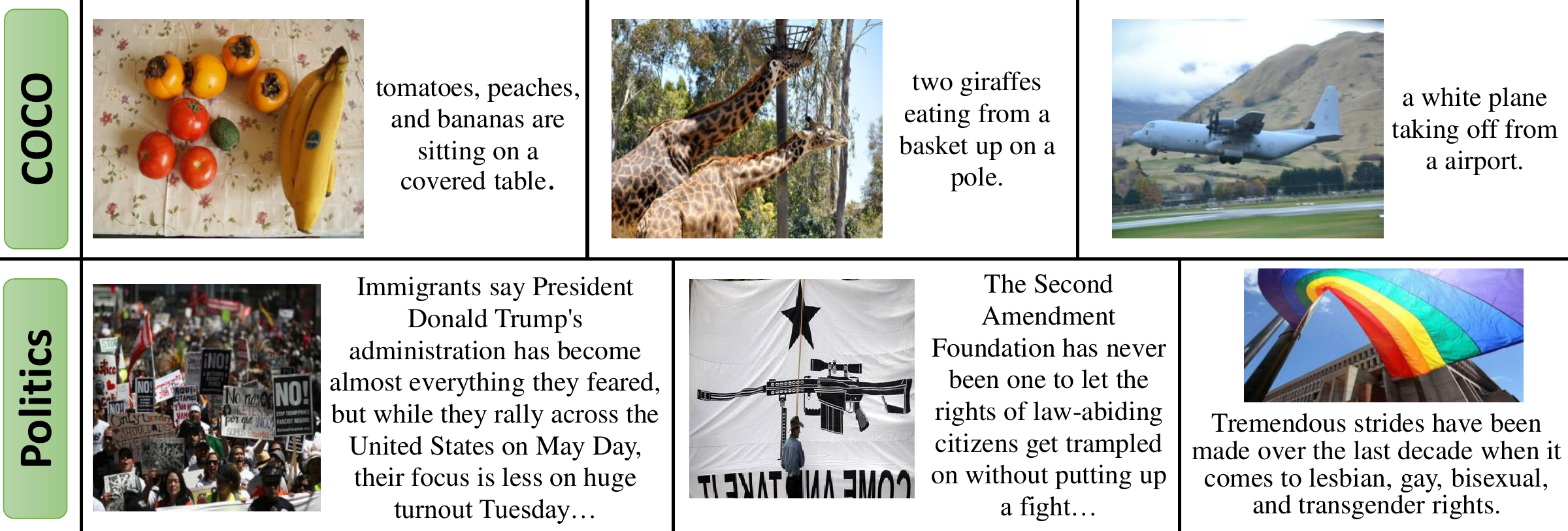}
    \caption{
    Image-text pairs from COCO \cite{lin2014microsoft} and Politics \cite{thomas2019predicting}. 
    Traditional image captions (top) are descriptive of the image, while we focus on the more challenging problem of aligning images and text with a non-literal complementary relationship (bottom). 
    }
    \label{fig:coco_vs_politics}
\end{figure*}

\begin{figure*}[t]
    \centering
    \includegraphics[width=1\textwidth]{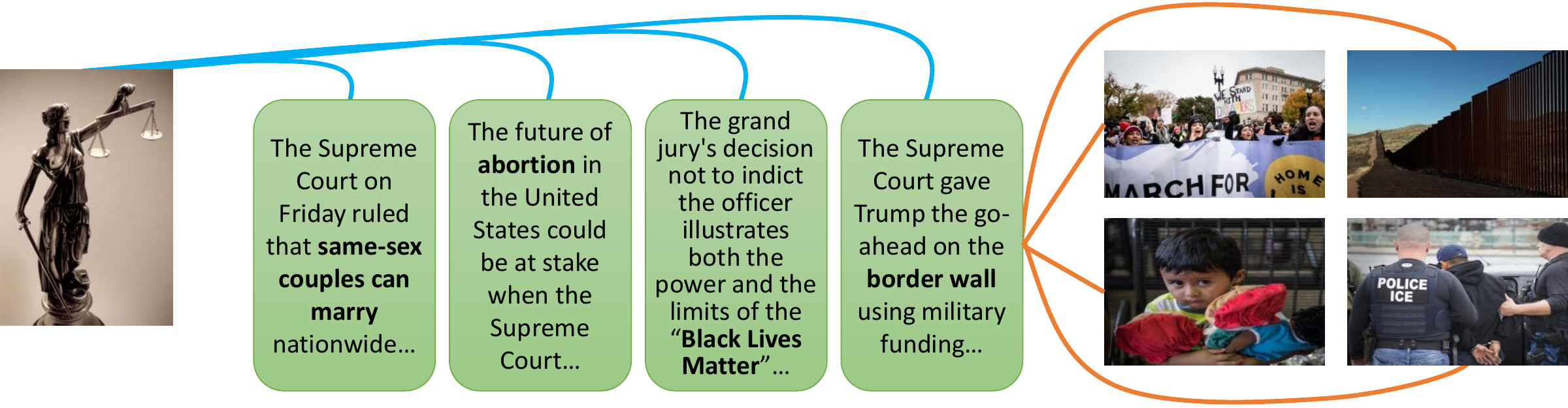}
    \caption{
    The image on the left symbolizes justice and may be paired with text about a variety of subjects (e.g.~abortion, same sex marriage). Similarly, text regarding immigration (right) may be paired with visually dissimilar images. Our approach enforces that \textit{semantically} similar content (images on the right) is close in the learned space. To discover such content, we use semantic neighbors of the text and their paired images.
    }
    \label{fig:complementarity}
\end{figure*}

Vision-language tasks such as image captioning \cite{you2016image, anderson2018bottom, lu2018neural} and cross-modal generation and retrieval \cite{reed2016generative, zhang2017stackgan, zhang2018photographic} have seen increased interest in recent years. 
At the core of methods in this space are techniques to bring together images and their corresponding pieces of text. However, most existing cross-modal retrieval methods only work on data where the two modalities (images and text) are well aligned, and provide fairly redundant information.
As shown in Fig.~\ref{fig:coco_vs_politics}, captioning datasets such as COCO contain samples where the overlap between images and text is significant (both image and text mention or show the same objects). 
In this setting, cross-modal retrieval means finding the manifestation of a single concept in two modalities (e.g. learning embeddings such that the word ``banana'' and the pixels for ``banana'' project close by in a learned space). 

In contrast, real-world news articles contain image and text pairs that cover the same topic, but show complementary information (protest signs vs information about the specific event; guns vs discussion of rights; rainbow flag vs LGBT rights).
While a human viewer can still guess which images go with which text, the alignment between image and text is abstract and symbolic.
Further, images in news articles are ambiguous \emph{in isolation}.
We show in Fig.~\ref{fig:complementarity} that an image might illustrate multiple related texts (shown in green), and each text in turn could be illustrated with multiple visually distant images (e.g. the four images on the right-hand side could appear with the border wall text).
Thus, we must first resolve any ambiguities in the image, and figure out ``what it means''.

We propose a metric learning approach where we use the semantic relationships between text segments, to guide the embedding learned for corresponding images.
In other words, to understand what an image ``means'', we look at what articles it appeared with.
Unlike prior approaches, we capture this information not only across modalities, but within the image modality itself. 
If texts $y_i$ and $y_j$ are semantically similar, we learn an embedding where we explicitly encourage their paired images $x_i$ and $x_j$ to be similar, using a new unimodal loss. 
Note that in general $x_i$ and $x_j$ need not be similar in the original visual space (Fig.~\ref{fig:complementarity}). 
In addition, we encourage texts $y_i$ and $y_j$, who were close in the unimodal space, to remain close. 

\begin{figure*}[t]
    \centering
    \includegraphics[width=1\textwidth]{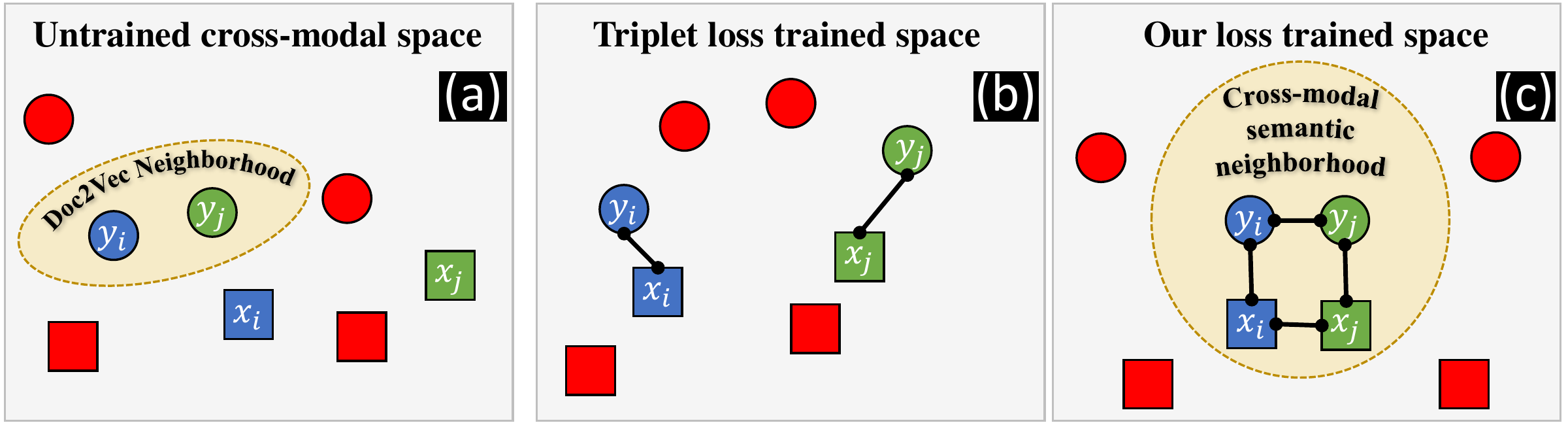}
    \caption{We show how our method enforces cross-modal semantic coherence. Circles represent text and squares images. In (a), we show the untrained cross-modal space. Note $y_i$ and $y_j$ are neighbors in Doc2Vec space and thus semantically similar. (b) shows the space after triplet loss training. $y_i$ and $x_i$, and $y_j$ and $x_j$, are now close as desired, but $y_i$ and $y_j$ have moved apart, and $x_i$ and $x_j$ remain distant. 
    (c) shows our loss's effect. Now, all semantic neighbors (both images and text) are pulled closer.}
    \label{fig:loss_effect_on_learned_space}
\end{figure*}

Our novel loss formulation explicitly encourages \textit{within-modality semantic coherence}. 
Fig.~\ref{fig:loss_effect_on_learned_space} shows the effect. On the left, we show the proximity of samples before cross-modal learning; specifically, while two texts are close in the document space, their paired articles may be far from the texts. In the middle, we show the effect of using a standard triplet loss, which pulls image-text pairs close, but does not necessarily preserve the similarity of related articles; they are now further than they used to be in the original space. In contrast, on the right, we show how our method brings paired images and text closer, while also preserving a semantically coherent region, i.e. the texts remained close.

In our approach, we use neighborhoods in the original text document space, to compute semantic proximity.
We also experiment with an alternative approach where we compute neighborhoods using the visual space, then guide the corresponding texts to be close. This approach is a variant of ours, and is novel in the sense that it uses proximity in one unimodal space, to guide the other space/modality. While unimodal losses based on visual similarity are helpful over a standard cross-modal loss (e.g. triplet loss), our main approach is superior.

Next, we compare to a method \cite{wang2016learning} which utilizes the \emph{set} of text annotations available for an image in COCO, to guide the structure of the learned space.
We show that when these ground-truth annotations are available, using them to compute neighborhoods in the textual space is the most reliable. However, on many datasets, such sets of annotations (more than one for the same image) are not available. We show that our approach offers a comparable alternative.

Finally, we test the contribution of our additional losses using PVSE \cite{song2019polysemous}, a state-of-the-art visual semantic embedding model, as a backbone. We show that our proposed loss further improves the performance of this model.

To summarize, our contributions are as follows. 
\begin{itemize}[nolistsep,noitemsep]
\item We preserve relationships in the original \textit{semantic} space. Because images do not clearly capture semantics, we use the semantic space (from text) to guide the image representation, through a unimodal (within-modality) loss. 
\item We perform detailed experimental analysis of our proposed loss function, including ablations, on four recent large-scale image-text datasets. One \cite{biten2019good} contains multimodal articles from New York Times, and another contains articles from far-left/right media \cite{thomas2019predicting}. We also conduct experiments on \cite{sharma2018conceptual,lin2014microsoft}. Our approach significantly improves the state-of-the-art in most cases. The more abstract the dataset/alignment, the more beneficial our approach.
\item We tackle a new cross-modal retrieval problem where the visual space is much less concrete. This scenario is quite practical, and has applications ranging from automatic caption generation for news images, to detection of fake multimodal articles (i.e. detecting whether an image supports the text). 
\end{itemize}
\section{Related Work}

\textit{Cross-modal learning.} 
A fundamental problem in cross-modal inference is the creation of a shared semantic manifold on which multiple modalities may be represented. The goal is to learn a space where content about related semantics (e.g. images of ``border wall'' and text about ``border wall'') projects close by, regardless of which modality it comes from. 
Many image-text embedding methods rely on a two-stream architecture, with one stream handling visual content (e.g. captured by a CNN) and the other stream handling textual content (e.g. through an RNN). Both streams are trained with paired data, e.g.~an image and its captions, and a variety of loss functions are used to encourage both streams to produce similar embeddings for paired data.  
Recently, purely attention-based approaches have been proposed \cite{lu2019vilbert,chen2019uniter}.
One common loss used to train retrieval models 
is triplet loss, which originates in the (single-modality) metric learning literature, e.g. for learning face representations \cite{schroff2015facenet}. In cross-modal retrieval, the triplet loss has been used broadly \cite{murrugarra2019cross,Zhu_2019_CVPR,Mithun_2019_CVPR,Pang_2019_CVPR,ye2018advise,faghri2017vse++}. 
Alternative choices include angular loss \cite{wang2017deep}, N-pairs loss \cite{sohn2016improved}, hierarchical loss \cite{ge2018deep}, and clustering loss \cite{oh2017deep}.

While single-modality losses like triplet, angular and N-pairs have been used across and within modalities, they are not sufficient for cross-modal retrieval. 
These losses do not ensure that the general semantics of the text are preserved in the new cross-modal space; thus, the cross-modal matching task might distort them too much. This phenomenon resembles forgetting \cite{lwf,goodfellow2013empirical} but in the cross-modal domain.
Our method preserves within-modal structure, and a similar effect can be achieved by leveraging category labels as in \cite{Zhen_2019_CVPR,wang2017adversarial,marin2019recipe1m+,carvalho2018cross}; however, such labels are not available in the datasets we consider, nor is it clear how to define them, since matches lie beyond the presence of objects.
Importantly, classic retrieval losses losses do not tackle the complementary relationship between images and text, which makes the space of topically related images more visually diffuse. In other words, two images might depict substantially \textit{different visual} content but nonetheless be \textit{semantically related}.

Note that we do not propose a new \emph{model} for image-text alignment, but instead propose cross-modal embedding \emph{constraints} which can be used to train any such model. For example, we compare to Song et al. \cite{song2019polysemous}'s recent polysemous visual semantic embedding (PVSE) model, which uses global and local features to compute self-attention residuals. Our loss improves \cite{song2019polysemous}'s performance. 

Our work is also related to cross-modal distillation \cite{frome2013devise, socher2013zero, Gupta_2016_CVPR, girdhar2019distinit}, 
which transfers supervision across modalities, but none of these approaches exploit the semantic signal that text neighborhoods carry to constrain the visual representations. Finally, \cite{Zhang_2018_BMVC,kruk2019integrating,alikhani2020clue} detect different types of image-text relationships (e.g. parallel, complementary) but do not retrieve across modalities.

\textit{Metric learning}
approaches learn distance metrics which meaningfully measure the similarity of objects. 
These can be broadly categorized into: 1) sampling-based methods \cite{han2015matchnet, simo2015discriminative, wang2015unsupervised, shrivastava2016training, yuan2017hard, lu2017discriminative, oh2017deep, harwood2017smart, wu2017sampling, lu2019sampling, wang2019multi}, which intelligently choose easy/hard samples or weight samples; or 2) loss functions \cite{hadsell2006dimensionality, weinberger2009distance, schroff2015facenet, sohn2016improved, wang2017deep, duan2018deep, ge2018deep} which impose intuitions regarding neighborhood structure, data separation, etc.
Our method relates to the second category. 
Triplet loss \cite{schroff2015facenet, hoffer2015deep} takes into account the \textit{relative} similarity of positives and negatives, such that positive pairs are closer to each other than positives are to negatives. \cite{zhang2016embedding} generalize triplet loss by fusing it with classification loss. \cite{oh2016deep} integrate all positive and negative pairs within a minibatch, such that all pair combinations are updated jointly. Similarly, \cite{sohn2016improved}'s N-pair loss pushes multiple negatives away in each triplet. \cite{wang2016learning} propose a structural loss, which pulls multiple text paired with the same image together, but requires more than one ground truth caption per image (which most datasets lack). In contrast, our approach pulls semantically similar images \textit{and} text together and only requires a single caption per image.
More recently, \cite{wang2017deep} propose an angular loss which leverages the triangle inequality to constrain the angle between points within triplets. We show how cross-modal complementary information (semantics paired with diverse visuals) can be leveraged to improve the learned embedding space, regardless of the specific loss used. 

\section{Method}
\label{sec:method}

Consider two image-text pairs, $\{x_i, y_i\}$ and $\{x_j, y_j\}$. To ground the ``meaning'' of the images, we use proximity in a generic, pre-trained textual space between the texts $y_i$ and $y_j$. If $y_i$ and $y_j$ are semantically close, we expect that they will also be relatively close in the learned space, and further, that $x_i$ and $x_j$ will be close also. 
We observed that, while intuitive, this expectation does not actually hold in the learned cross-modal space. The problem becomes more severe when image and paired text do not exhibit literal alignment, as shown in Fig.~\ref{fig:coco_vs_politics}, because images paired via text neighbors could be visually different. 

\begin{figure*}[t]
    \centering
    \includegraphics[width=1\textwidth]{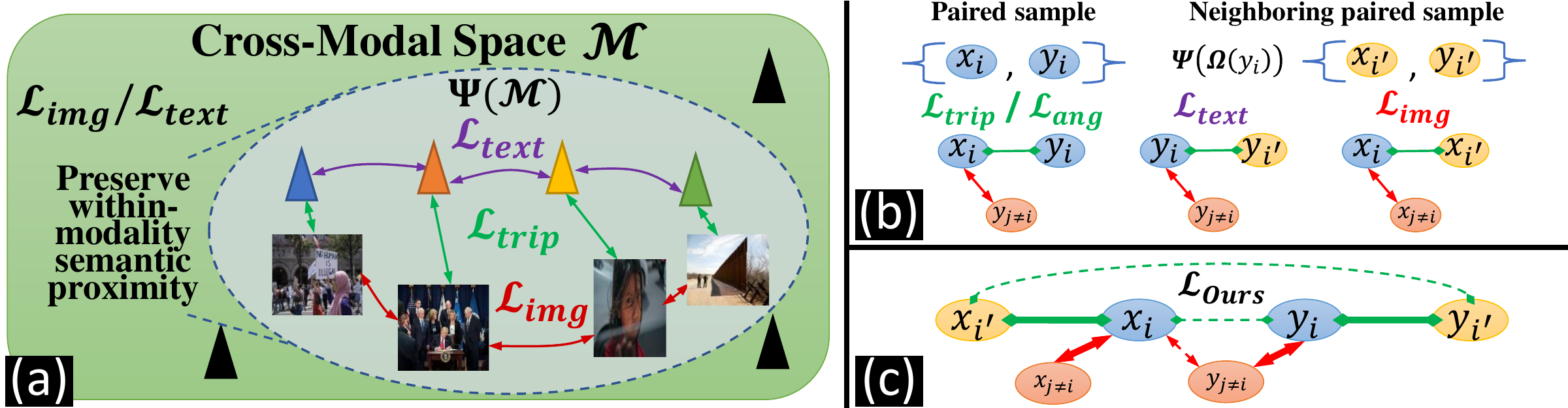}
    \caption{(a): $\mathcal{L}_{text}$ and $\mathcal{L}_{img}$ pull semantic neighbors of the same modality closer. The images are visually distinct, but semantically similar. (b): Pull connections are shown in green, and push in red. $\mathcal{L}_{trip}$ and $\mathcal{L}_{ang}$ operate cross-modally, but impose no within-modality constraints. (c): $\mathcal{L}_{ours}$ (which combines all three losses above) exploits the paired nature of the data to enforce the expected inter/intra-modal relationships. Solid lines indicate connections that our loss enforces but triplet/angular do not.
    }
    \label{fig:combined}
\end{figure*}

We describe how several common existing loss functions tackle cross-modal retrieval, and discuss their limitations. We then propose two constraints which pull within-modality semantic neighbors close to each other. Fig.~\ref{fig:combined} illustrates how our approach differs from standard metric learning losses. 

\subsection{Problem formulation and existing approaches}

We assume a dataset $\mathbfcal{D} = \left\{\mathbf{I}, \mathbf{T} \right\}$ of $n$ image-text pairs, where $\textbf{I} = \left\{ x_1,x_2,\ldots,x_n\right\}$ and $\textbf{T} = \left\{ y_1,y_2,\ldots,y_n\right\}$ denote the set of paired images and text, respectively. By pairs, we mean $y_i$ is text related to or co-occurring with image $x_i$. 
Let $f_I$ denote a convolutional neural network which projects images into the joint space and $f_T$ a recurrent network which projects text. We use the notational shorthand $f_T\left(y\right)=y$ and $f_I\left(x\right)=x$. The goal of training $f_I$ and $f_T$ is to learn a cross-modal manifold $\mathcal{M}$ where semantically similar samples are close. 
At inference time, we wish to retrieve a ground-truth paired text given an input image, or vice versa.
One common technique is triplet loss \cite{schroff2015facenet} which posits that
paired samples should be closer to one another than they are to non-paired samples. Let $\mathcal{T} = \left(x_i^a, y_i^p, y_j^n \right)$ denote a triplet of samples consisting of an anchor ($a$), positive or paired sample ($p$), and negative or non-paired sample ($n$) chosen randomly such that $i \neq j$. Let $m$ denote a margin. The triplet loss $\mathcal{L}_{trip}$ is then:
\begin{equation}
\label{eq:triplet_loss}
\mathcal{L}_{trip}\left(\mathcal{T}\right) = \left[ \norm{x_i^a - y_i^p}_2^2 - \norm{x_i^a - y_j^n}_2^2 + m \right]_+ 
\end{equation}

This loss is perhaps the most common one used in cross-modal retrieval tasks, but it has some deficiencies. For example, the gradient of the triplet loss wrt.~each point only considers two points, but ignores their relationship with the third one; for example, $\frac{\partial \mathcal{L}_{trip}}{\partial x_i^a} = 2\left(y_j^n-y_i^p\right)$. This allows for degenerate cases, so angular loss $\mathcal{L}_{ang}$ \cite{wang2017deep} accounts for the angular relationship of all three points:
\begin{equation}
\label{eq:angular_loss}
\mathcal{L}_{ang} \left(\mathcal{T}\right) = \left[ \norm{x_i^a - y_i^p}_2^2 - 4 \tan^2 \alpha \norm{y_j^n - \mathcal{C}_i}_2^2 \right]_+ 
\end{equation}
where $\mathcal{C}_i = \left( x_i^a + y_i^p \right) / 2$ is the center of a circle through anchor and positive. 

One challenging aspect of these losses is choosing a good negative term in the triplet. If the negative is too far from the anchor, the loss becomes 0 and no learning occurs. In contrast, if negatives are chosen too close, the model may have difficulty converging to a reasonable solution as it continuously tries to move samples to avoid overlap with the negatives. 
How to best sample triplets to avoid these issues is an active area of research \cite{duan2018deep}. One recent technique, the N-pairs loss \cite{sohn2016improved}, proposes that instead of a single negative sample being used, all negatives within the minibatch should be used. The N-pairs loss $\mathcal{L}_{ang}^{NP}$ pushes the anchor and positive embedding away from \textit{multiple} negatives simultaneously:
\begin{equation}
\label{eq:npairs_triplet_loss}
\mathcal{L}_{ang}^{NP} \left(\mathcal{T}\right) = \sum_{y_j \in \mathrm{minibatch}, ~ j \neq i} \mathcal{L}_{ang}\left(x_i^a, y_i^p, y_j^n \right) 
\end{equation}

The symmetric constraint \cite{zhou2017point} can also be added to explicitly account for bidirectional retrieval, i.e.~text-to-image, by swapping the role of images and text to form symmetric triplets $\mathcal{T}_{sym} = \left(y_i^a, x_i^p, x_i^n \right)$:
\begin{equation}
\label{eq:npairs_symmetric_loss}
\mathcal{L}_{ang}^{NP+SYM} \left(\mathcal{T}, \mathcal{T}_{sym}\right) = \mathcal{L}_{ang}^{NP}\left(\mathcal{T}\right) + \mathcal{L}_{ang}^{NP}\left(\mathcal{T}_{sym}\right) 
\end{equation}

\paragraph{Limitations.}
While these loss functions have been used for cross-modal retrieval, they 
do not take advantage of several unique aspects of the multi-modal setting. 
Only the dashed pull/push connections in Fig.~\ref{fig:combined} (c) are part of  triplet/angular loss. The solid connections are intuitive, but only enforced in our novel formulation.
We argue the lack of explicit \textit{within-modality} constraints allows discontinuities within the space for semantically related content from the same modality.

\subsection{Our proposed loss}
\label{sec:method:semantic_locality}

The text domain provides a semantic fingerprint for the image-text pair, since vastly dissimilar visual content may still be semantically related (e.g.~image of White house, image of protest), while similar visual content (e.g.~crowd in church, crowd at mall) could be semantically unrelated. We thus use the text domain to constrain within-modality semantic locality for both images and text.

To measure ground-truth semantic similarity, we pretrain a Doc2Vec \cite{le2014distributed} model $\Omega$ on the train set of text. Specifically, let $d$ be the document embedding of article $y_i$, $T$ denote the number of words in $y_i$, 
$w_{t}$ represent the embedding learned for word $t$, $p(\cdot)$ be the probability of the given word, and $k$ denote the look-around window. $\Omega$ learns word embeddings and document embeddings which maximize the average log probability: $\frac{1}{T} \sum_{t=1}^{T} \log p\left(w_{t} | d, w_{t-k}, \ldots, w_{t+k} \right) $. After training $\Omega$, we use iterative backpropagation to compute the document embedding which maximizes the log probability for every article in the dataset: $\Omega(\mathbf{T})=\left\{\Omega\left(y_1\right),\ldots,\Omega\left(y_n\right)\right\}$. 

Because Doc2Vec has been shown to capture latent topics within text documents well \cite{niu2015topic2vec}, we seek to enforce that locality originally captured in $\Omega(\mathbf{T})$'s space also be preserved in the cross-modal space $\mathcal{M}$. Let
\begin{equation}
    \Psi\left(\Omega(y_i)\right) = \left< x_{i^\prime}, y_{i^\prime}\right>
    \label{eq:neighbors}
\end{equation}
denote a nearest neighbor function over $\Omega(\mathbf{T})$, where $\left< \cdot, \cdot\right>$ is an image-text pair in the train set randomly sampled from the $k=200$ nearest neighbors to $y_i$, and $i \neq i^\prime$. $\Psi\left(\Omega(y_i)\right)$ thus returns an image-text pair semantically related to $y_i$.

We formulate two loss functions to enforce within-modality semantic locality in $\mathcal{M}$. The first, $\mathcal{L}_{text}$, enforces locality of the text's projections: 
\begin{equation}
\label{eq:text_loss}
\begin{gathered}
\mathcal{T}_{text}^\prime = \left(y_i^a, y_{i^\prime}^p, y_j^n \right)  \\
\mathcal{L}_{text}\left(\mathcal{T}_{text}^\prime\right) = \mathcal{L}_{ang} \left(\mathcal{T}_{text}^\prime\right) \\ 
\mathcal{L}_{ang} \left(\mathcal{T}_{text}^\prime\right) = \left[ \norm{y_i^a - y_{i^\prime}^p}_2^2 - 4 \tan^2 \alpha \norm{y_j^n - \mathcal{C}_i}_2^2 \right]_+ 
\end{gathered}
\end{equation}
where $y_j^n$ is the negative sample chosen randomly such that $i \neq j$ and $\mathcal{C}_i = \left( y_i^a + y_i^p \right) / 2$. $\mathcal{L}_{text}$ is the most straightforward transfer of semantics from $\Omega(\mathbf{T})$'s space to the joint space: nearest neighbors in $\Omega$ should remain close in $\mathcal{M}$. 

As Fig.~\ref{fig:combined} (c) shows, $\mathcal{L}_{text}$ also indirectly causes semantically related images to move closer in $\mathcal{M}$: there is now a weak connection between $x_i$ and $x_{i^\prime}$ through the now-connected $y_i$ and $y_{i^\prime}$.
To directly ensure smoothness and semantic coherence between $x_i$ and $x_{i^\prime}$, 
we propose a second constraint, $\mathcal{L}_{img}$:
\begin{equation}
\label{eq:image_loss}
\begin{gathered}
\mathcal{T}_{img}^\prime = \left(x_i^a, x_{i^\prime}^p, x_j^n \right)   \\ 
\mathcal{L}_{img}\left(\mathcal{T}_{img}^\prime\right) = \mathcal{L}_{ang} \left(\mathcal{T}_{img}^\prime\right) \\ 
\mathcal{L}_{ang} \left(\mathcal{T}_{img}^\prime\right) = \left[ \norm{x_i^a - x_{i^\prime}^p}_2^2 - 4 \tan^2 \alpha \norm{x_j^n - \mathcal{C}_i}_2^2 \right]_+ 
\end{gathered}
\end{equation}
where $x_j^n$ is the randomly chosen negative sample such that $i \neq j$ and $\mathcal{C}_i = \left( x_i^a + x_i^p \right) / 2$. Note that $x_i$ and $x_{i^\prime}$ are often not going to be neighbors in the original visual space.
We use N-pairs over all terms to maximize discriminativity, and symmetric loss to ensure robust bidirectional retrieval:
\begin{gather}
\mathcal{L}_{ang}^{OURS}\left(\mathcal{T}, \mathcal{T}_{sym}, \mathcal{T}_{text}^\prime, \mathcal{T}_{img}^\prime \right) = \\ 
\mathcal{L}_{ang}^{NP+SYM} \left(\mathcal{T}, \mathcal{T}_{sym}\right) + 
\alpha \mathcal{L}_{text}^{NP}\left(\mathcal{T}_{text}^\prime\right) + \beta \mathcal{L}_{img}^{NP}\left( \mathcal{T}_{img}^\prime \right) \notag
\end{gather}
where $\alpha, \beta$ are hyperparameters controlling the importance of each constraint. 

\vspace{-0.3cm}
\paragraph{Second variant.}
We also experiment with a variant of our method where the nearest neighbor function in Eq.~\ref{eq:neighbors} (computed in Doc2Vec space) is replaced with one that computes nearest neighbors in the space of visual (e.g. ResNet) features. Now $x_i, x_{i^\prime}$ are neighbors in the original visual space before cross-modal training, and $y_i, y_{i^\prime}$ are their paired articles (which may not be neighbors in the original Doc2Vec space). We denote this method as \textsc{Ours} (Img NNs) in Table \ref{tab:main_result}, and show that while it helps over a simple triplet- or angular-based baseline, it is inferior to our main method variant described above.

\vspace{-0.3cm}
\paragraph{Discussion.}
At a low level, our method combines three angular losses. However, note that our losses in Eq.~\ref{eq:text_loss} and Eq.~\ref{eq:image_loss} do not exist in  prior literature. While \cite{wang2016learning} leverages ground-truth neighbors (sets of neighbors provided together for the same image sample in a dataset), we are not aware of prior work that estimates neighbors. Importantly, we are not aware of prior work that uses the text space to construct a loss over the image space, as Eq.~\ref{eq:image_loss} does.
We show that the choice of space in which semantic coherency is computed is important; doing this in the original textual space is superior than using the original image space.
We show the contribution of both of these losses in our experiments.

\subsection{Implementation details}

All methods use a two-stream architecture, with the image stream using a ResNet-50 \cite{he2016deep} architecture initialized with ImageNet features, and the text stream using Gated Recurrent Units \cite{cho2014properties} with hidden state size 512. We use image size 224x224 and random horizontal flipping, and initialize all non-pretrained learnable weights via Xavier init.~\cite{glorot2010understanding}. Text models are initialized with word embeddings of size 200 learned on the target dataset. We apply a linear transformation to each model's output features ($\mathbb{R}^{2048\times256}$ for image, $\mathbb{R}^{512\times256}$ for text) to get the final embedding, and perform $L_2$ normalization.
We use Adam \cite{kingma2014adam} with minibatch size 64, learning rate 1.0e-4, and weight decay 1e-5. We decay the learning rate by a factor of 0.1 after every 5 epochs of no decrease in val.~loss. We use a train-val-test split of 80-10-10. 
For Doc2Vec, we use \cite{rehurek_lrec} with $d \in \mathbb{R}^{200}$ and train using distributed memory \cite{le2014distributed} for 20 epochs with window $k=20$, ignoring words that appear less than 20 times. We use hierarchical softmax \cite{morin2005hierarchical} to compute $p(\cdot)$. To efficiently compute approximate nearest neighbors for $\Psi$, we use \cite{malkov2016efficient}; our method adds negligible computational overhead as neighbors are computed prior to training. 
We choose $\alpha=0.3, \beta=0.1$ for  $\mathcal{L}_{trip}^{OURS}$, and $\alpha=0.2, \beta=0.3$ for  $\mathcal{L}_{ang}^{OURS}$, on a held-out val.~set.
\section{Experiments}

We compare our method to five baselines on four recent large-scale datasets. Our results consistently demonstrate the superiority of our approach at bidirectional retrieval. We also show our method better preserves within-modality semantic locality by keeping neighboring images and text closer in the joint space. 

\subsection{Datasets}
Two datasets feature challenging indirect relations between image and text, compared to standard captioning data. These also exhibit longer text paired with images: 59 and 18 words on average, compared to 11 in COCO. 

\textbf{Politics} \cite{thomas2019predicting} 
consists of images paired with news articles. In some cases, multiple images were paired with boilerplate text (website headliner, privacy policy) due to failed data scraping. We removed duplicates using MinHash \cite{broder1997resemblance}. We were left with 246,131 unique image-text pairs. Because the articles are lengthy, we only use the first two sentences of each.
\cite{thomas2019predicting} do not perform retrieval.

\textbf{GoodNews} \cite{biten2019good} consists of $\sim$466k images paired with their captions. All data was harvested from the New York Times. Captions often feature abstract or indirect text in order to relate the image to the article it appeared with. The method in \cite{biten2019good} takes image and text as input, hence cannot serve as a baseline.

We also test on two large-scale standard image captioning datasets, where the relationship between image and text is typically more direct:

\textbf{COCO} \cite{lin2014microsoft} is a large dataset containing numerous annotations, such as objects, segmentations, and captions. The dataset contains $\sim$120k images with captions. Unlike our other datasets, COCO contains more than one caption per image, with each image paired with four to seven captions.

\textbf{Conceptual Captions} \cite{sharma2018conceptual} is composed of $\sim$3.3M image-text pairs. The text comes from automatically cleaned alt-text descriptions paired with images harvested from the internet and has been found to represent a much wider variety of style and content compared to COCO. 

\subsection{Baselines}
We compare to N-Pairs Symmetric Angular Loss (\textsc{Ang+NP+Sym}, a combination of \cite{wang2017deep,sohn2016improved,zhou2017point}, trained with $\mathcal{L}_{ang}^{NP+SYM}$). For a subset of results, we also replace the angular loss with the weaker but more common triplet loss (\textsc{Trip+NP+Sym}).
We show the result of choosing to enforce coherency within the image and text modalities by using images rather than text; this is the second variant of our method, denoted \textsc{Ours} (Img NNs).

We also compare our approach against the deep structure preserving loss \cite{wang2016learning} (\textsc{Struc}), which enforces that captions paired with the same image are closer to each other than to non-paired captions. 

Finally, we show how our approach can improve the performance of a state-of-the-art cross-modal retrieval model. \textsc{PVSE} \cite{song2019polysemous} uses both images and text to compute a self-attention residual before producing embeddings.

\subsection{Quantitative results}

We formulate a cross-modal retrieval task such that given a query image or text, the embedding of the paired text/image must be closer to the query embedding than non-paired samples also of the target modality. We sample random (non-paired) samples from the test set, along with the ground-truth paired sample. We then compute Recall@1 within each task: that is, whether the ground truth paired sample is closer to its cross-modal embedding than the non-paired embeddings. For our most challenging datasets (GoodNews and Politics), we use a 5-way task. For COCO and Conceptual Captions, we found this task to be too simple and that all methods easily achieved very high performance due to the literal image-text relationship. Because we wish to distinguish meaningful performance differences between methods, we used a 20-way task for Conceptual Captions and a 100-way task for COCO. Task complexities were chosen based on the baseline's performance, before our method's results were computed. 

\begin{table}[t]
\centering
\begin{tabular}{c|c|c|c|c||c|c|c|c|}
\cline{2-9}
 & \multicolumn{4}{c||}{\textbf{Img-Text Non-Literal}} & \multicolumn{4}{c|}{\textbf{Img-Text Literal}} \\ 
\cline{2-9}
 & \multicolumn{2}{c|}{\textbf{Politics} \cite{thomas2019predicting}} & \multicolumn{2}{c||}{\textbf{GoodNews} \cite{biten2019good}} & \multicolumn{2}{c|}{\textbf{ConcCap} \cite{sharma2018conceptual}} & \multicolumn{2}{c|}{\textbf{COCO} \cite{lin2014microsoft}} \\ \hline
\multicolumn{1}{|c||}{\textbf{Method}} & \textbf{I$\rightarrow$T} & \textbf{T$\rightarrow$I} &  \textbf{I$\rightarrow$T} & \textbf{T$\rightarrow$I} & \textbf{I$\rightarrow$T} & \textbf{T$\rightarrow$I} & \textbf{I$\rightarrow$T} & \textbf{T$\rightarrow$I} \\ \hline \hline
\multicolumn{1}{|c||}{\textsc{Ang+NP+Sym}} & 0.6270 & 0.6216 & 0.8704 & 0.8728 & 0.7687 & 0.7695 & \textbf{0.6976} & \textbf{0.6964} \\ \hline
\multicolumn{1}{|c||}{\textsc{Ours} (Img NNs)} & 0.6370 & 0.6378 & 0.8840 & 0.8852 & 0.7636 & 0.7666 & 0.6819 & 0.6876\\ \hline 
\multicolumn{1}{|c||}{\textsc{Ours} } & \textbf{0.6467} & \textbf{0.6492} & \textbf{0.8849} & \textbf{0.8865} & \textbf{0.7760} & \textbf{0.7835} & 0.6900 & 0.6885 \\ \hline \hline
\multicolumn{1}{|c||}{\textsc{PVSE}} & 0.6246 & 0.6199 & 0.8724 & 0.8709 & 0.7746 & 0.7809 & 0.6878 & 0.6892 \\ \hline 
\multicolumn{1}{|c||}{\textsc{PVSE}+Ours} & \textbf{0.6264} & \textbf{0.6314} & \textbf{0.8867} & \textbf{0.8864} & \textbf{0.7865} & \textbf{0.7924} & \textbf{0.6932} & \textbf{0.6925} \\ \hline
\hline
\multicolumn{1}{|c||}{\textsc{Trip+NP+Sym}} & 0.4742 & 0.4801 & 0.7203 & 0.7216 & \textbf{0.5413} & 0.5332 & \textbf{0.4957} & \textbf{0.4746} \\ \hline
\multicolumn{1}{|c||}{\textsc{Ours (Trip)} } & \textbf{0.4940} & \textbf{0.4877} & \textbf{0.7390} & \textbf{0.7378} & 0.5386 & \textbf{0.5394} & 0.4790 & 0.4611 \\ \hline 
\end{tabular}
\caption{We show retrieval results for image to text (\textbf{I}$\rightarrow$\textbf{T}) and text to image (\textbf{T}$\rightarrow$\textbf{I}) on all datasets. The best method per group is shown in bold.}
\label{tab:main_result}
\end{table}

We report the results in Table \ref{tab:main_result}. 
The first and second group of results all use angular loss, while the third set use triplet loss. We observe that our method significantly outperforms all baselines tested for both directions of cross-modal retrieval for three of the four datasets. Our method achieves a 2\% relative boost in accuracy (on average across both retrieval tasks) vs.~the strongest baseline on GoodNews, and a 4\% boost on Politics. We also observe recall is much worse for all tasks on the Politics dataset compared to GoodNews, likely because the images and article text are much less well-aligned. The performance gap seems small but note that given the figurative use of images in these datasets, often there may not be a clear ground-truth answer. In Fig.~\ref{fig:complementarity}, Themis may be constrained to be close to protestors or border wall. At test time, the ground-truth text paired with Themis may be about the Supreme Court, but one of the ``incorrect'' answers could be about immigration or freedom, which still make sense. Our method keeps more neighbors closer to the query point as shown next, thus may retrieve plausible, but technically ``incorrect'' neighbors for a query.

Importantly, we see that while the variant of our method using neighborhoods computed in image space (\textsc{Ours} Img NNs) does outperform \textsc{Ang+NP+Sym}, it is weaker than our main method variant (\textsc{Ours}).
We also observe that when adding our loss on top of the PVSE model \cite{song2019polysemous}, accuracy of retrieval improves. In other words, our loss is complementary to advancements accomplished by network model-based techniques such as attention. 

Our method outperforms the baselines on ConcCap also, but not on COCO, since COCO is the easiest, least abstract of all datasets, with the most literal image-text alignment. Our approach constrains neighboring texts and their images to be close, and for datasets where matching is on a more abstract, challenging level, the benefit of neighbor information outweighs the disadvantage of this inexact similarity. However, for more straightforward tasks (e.g. in COCO), it may introduce noise. For example, for caption ``a man on a bicycle with a banana'', the model may pull that image and text closer to images with a banana in a bowl of fruit. 
Overall, our approach of enforcing within-modality semantic neighborhoods substantially improves cross-view retrieval, particularly when the relationship between image and text is complementary, rather than redundant.

To better ground our method's performance in datasets typically used for retrieval, we also conducted an experimented on Flickr30K \cite{plummer2015flickr30k}. Since that dataset does not exhibit image-text complementarity, we do not expect our method to improve performance, but it should not significantly reduce it. We compared the original PVSE against PVSE with our novel loss. We observed that our method slightly outperformed the original PVSE, on both text-to-image and image-to-text retrieval (0.5419 and 0.5559 for ours, vs 0.5405 and 0.5539 for PVSE).

\begin{table}[t]
\centering
\begin{tabular}{cc|c|c|c|c|}
\cline{3-6}
& & \textsc{Ours (Trip)} & \textsc{Struc} (GT, Text) & \textsc{Struc} (NN$_{\Omega}$, Text) & \textsc{Struc} (NN$_{\Omega}$, Img) \\ \hline
\multicolumn{1}{|c}{\multirow{2}{*}{\parbox{3.5em}{\centering \textbf{COCO}}}}
& \multicolumn{1}{|c||}{\textbf{I$\rightarrow$T}} & \underline{0.4790} & \textbf{0.4817} & 0.4635 & 0.4752 \\ 
\cline{2-6}
\multicolumn{1}{|c}{} & \multicolumn{1}{|c||}{\textbf{T$\rightarrow$I}} & \underline{0.4611} & \textbf{0.4867} & 0.4594 & 0.4604 \\ \hline
\end{tabular}
\caption{We show retrieval results for image to text (\textbf{I}$\rightarrow$\textbf{T}) and text to image (\textbf{T}$\rightarrow$\textbf{I}) on COCO using \cite{wang2016learning}'s loss vs.~ours. GT requires multiple \textbf{G}round \textbf{T}ruth captions per image, while NN uses \textbf{N}earest \textbf{N}eighbors. The best method per row is shown in bold, while the best method which does not require a set of neighboring text is underlined. }
\label{tab:deep_structure}
\end{table}

\begin{table}[t]
\centering
\vspace{-0.5em}
\begin{tabular}{c|c|c|c|c|}
\cline{2-5}
 & \multicolumn{2}{c|}{\textbf{GoodNews} \cite{biten2019good}} & \multicolumn{2}{c|}{\textbf{Politics} \cite{thomas2019predicting}} \\ \hline
\multicolumn{1}{|c|}{\textbf{Method}} & \textbf{I} & \textbf{T} &  \textbf{I} & \textbf{T} \\ \hline \hline
\multicolumn{1}{|c|}{\textsc{Trip+NP+Sym}} & 0.1183 & 0.1294 & 0.1135 & 0.1311 \\ \hline
\multicolumn{1}{|c|}{\textsc{Ours (Trip)}} & \textbf{0.1327} & \textbf{0.1426} & \textbf{0.1319} & \textbf{0.1483} \\ \hline \hline
\multicolumn{1}{|c|}{\textsc{Ang+NP+Sym}} & 0.1032 & 0.1131 & 0.1199 & 0.1544 \\ \hline
\multicolumn{1}{|c|}{\textsc{Ours (Ang)} } & \textbf{0.1270} & \textbf{0.1376} & \textbf{0.1386} & \textbf{0.1703} \\ \hline
\end{tabular}
\caption{We test how well each method preserves the semantic neighborhood (see text) of $\Omega$ in $\mathcal{M}$.  Higher values are better.  Best method is shown in bold. }
\vspace{-1.5em}
\label{tab:semantic_consistency}
\end{table}
\begin{table*}[t]
\centering
\begin{tabular}{c|c|c|c|c|}
\cline{2-5}
 & \multicolumn{2}{c|}{\textbf{GoodNews} \cite{biten2019good}} & \multicolumn{2}{c|}{\textbf{Politics} \cite{thomas2019predicting}} \\ \hline
\multicolumn{1}{|c|}{\textbf{Method}} & \textbf{I$\rightarrow$T} & \textbf{T$\rightarrow$I} &  \textbf{I$\rightarrow$T} & \textbf{T$\rightarrow$I} \\ \hline \hline
\multicolumn{1}{|c|}{\textsc{Ours (Ang)} } & \textbf{0.8849} & \textbf{0.8865} & \textbf{0.6467} & \textbf{0.6492} \\ \hline
\multicolumn{1}{|c|}{\textsc{Ours (Ang)}-$\mathcal{L}_{text}$ } & \underline{0.8786} & 0.8813 & 0.6387 &  \underline{0.6467} \\ \hline
\multicolumn{1}{|c|}{\textsc{Ours (Ang)}-$\mathcal{L}_{img}$ } & 0.8782 & \underline{0.8817} & \underline{0.6390} & 0.6413 \\ \hline
\end{tabular}
\caption{We show an ablation of our method where we remove either component of our loss.  The best method is shown in \textbf{bold} and the best ablation is underlined.}
\vspace{-0.5em}
\label{tab:ablation}
\end{table*}

\begin{figure*}[t]
    \centering
    \includegraphics[width=1\textwidth]{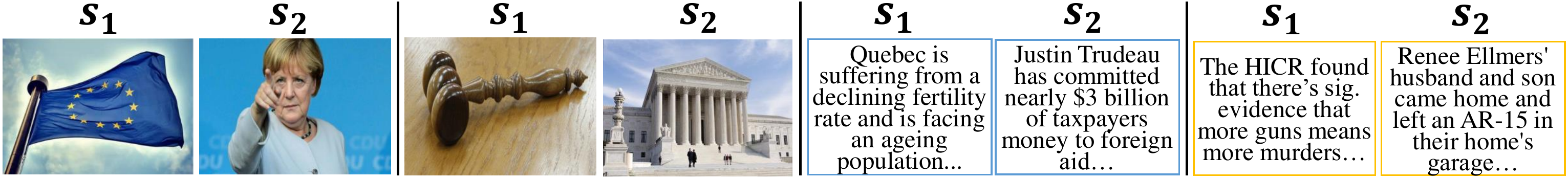}
    \caption{Uncurated results showing image/text samples that our method keeps closest in $\mathcal{M}$ compared to the baseline, i.e.~pairs where $\frac{d_{ours}\left(s_1,s_2\right)}{d_{baseline}\left(s_1,s_2\right)}$ is smallest.
    Our method keeps semantically related images and text closer in the space, relative to the baseline. While the images are not visually similar, they are semantically similar (EU and Merkel; judge's gavel and Supreme Court). 
    }
    \label{fig:relative_distance}
\end{figure*}

\begin{figure*}[t]
    \centering
    \includegraphics[width=1\textwidth]{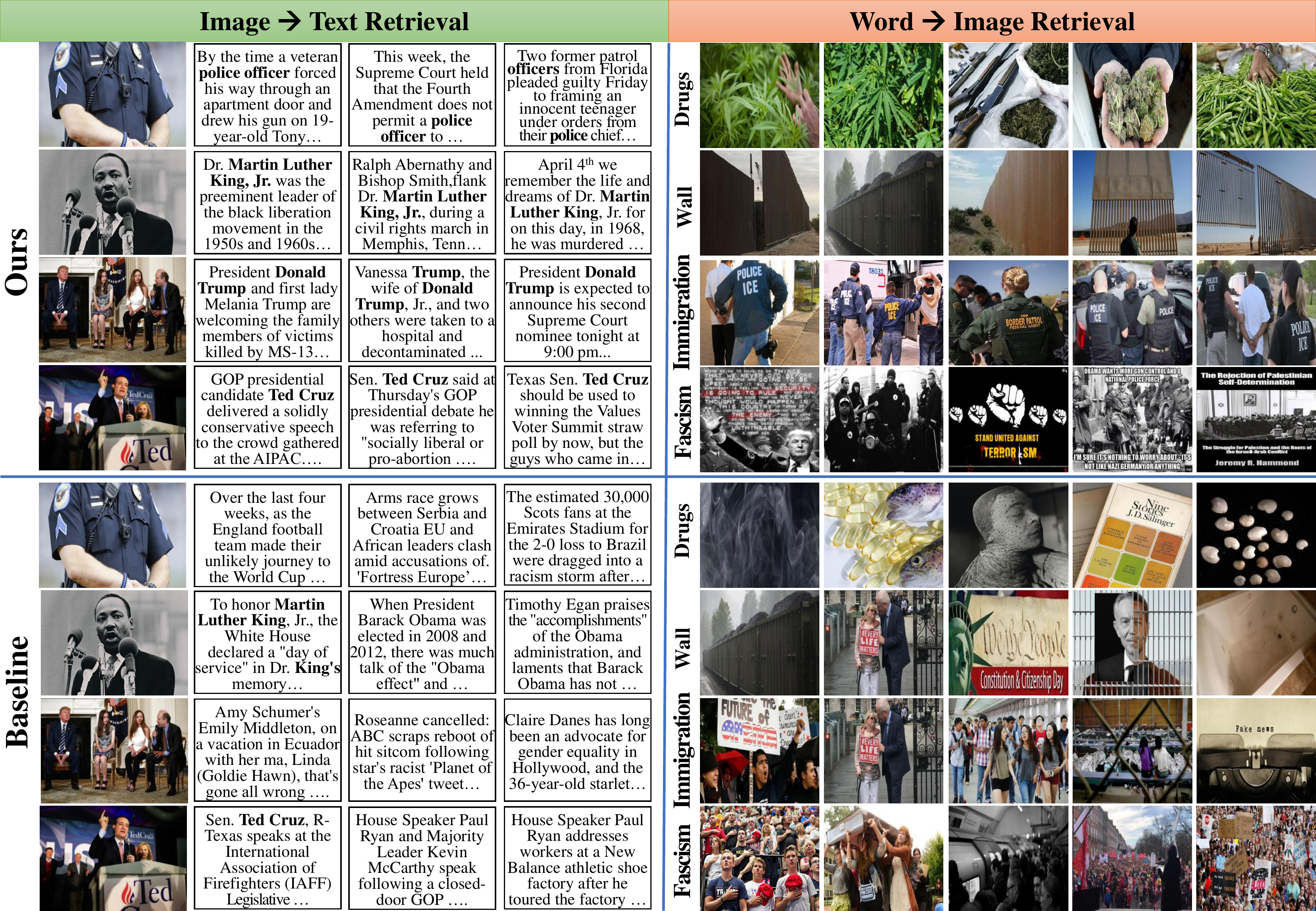}
    \caption{We show cross-modal retrieval results on Politics \cite{thomas2019predicting} using our method and the strongest baseline. We bold text aligning with the image. For text retrieval, ours returns more relevant (and semantically consistent) results. For image retrieval, our method exhibits more consistency (e.g.~drug images are marijuana, immigration images show arrests), while the baseline returns more inconsistent and irrelevant images. }
    \label{fig:cross_modal_retrieval}
\end{figure*}

In Table \ref{tab:deep_structure}, we show a result comparing our method to Deep Structure Preserving Loss \cite{wang2016learning}. Since this method requires a \textit{set} of annotations (captions) for an image, i.e. it requires \emph{ground-truth neighbor relations} for texts, we can only apply it on COCO. In the first column, we show our method. In the second, we show \cite{wang2016learning} using ground-truth neighbors.
Next, we show using \cite{wang2016learning} with \emph{estimated neighbors}, as in our method. We see that as expected, using estimated rather than ground-truth text neighbors reduces performance (third vs.~second columns). When estimated neighbors are used in \cite{wang2016learning}'s structural constraint, our method performs better (third vs.~first columns). Interestingly, we observe that defining \cite{wang2016learning}'s structural constraint in image rather than text space is better (fourth vs.~third columns). In both cases, neighborhoods are computed in \emph{text} space (Eq.~\ref{eq:neighbors}). This may be because the structural constraint, which requires the group of neighbors to be closer together than to others, is too strict for estimated text neighbors. That is, the constraint may require the text embeddings to lose useful discriminativity to be closer to neighboring text. Neighboring images are likely to be much more visually similar in COCO than in GoodNews or Politics as they will contain the same objects. 

We next test how well each method preserves the \textit{semantic neighborhood} given by $\Omega$, i.e.~Doc2Vec space. We begin by computing the embeddings in $\mathcal{M}$ (cross-modal space) for all test samples. For each such sample $s_i$ (either image or text), we compute $\Psi_{\mathcal{M}}\left(s_i\right)$, that is, we retrieve the neighbors (of the same modality as $s_i) $ in $\mathcal{M}$. We next retrieve the neighbors of $s_i$ in $\Omega$, $\Psi_\Omega\left(s_i\right)$,  described in Sec.~\ref{sec:method:semantic_locality}. For each sample, we compute $\lvert \Psi_{\mathcal{M}}\left(s_i\right) \cap \Psi_{\Omega}\left(s_i\right) \rvert \, / \, \lvert \Psi_{\Omega}\left(s_i\right) \rvert $, i.e.~the percentage of the nearest neighbors of the sample in $\Omega$ which are also its neighbors in $\mathcal{M}$.
That is, we measure how well each method preserves within-modality semantic locality through the number of neighbors in Doc2Vec space which remain neighbors in the learned space. 
We consider the 200 nearest neighbors.
We report the result for competitive baselines in Table $\ref{tab:semantic_consistency}$. 
We find that our constraints are, indeed, preserving within-modality semantic locality, as sample proximity in $\Omega$ is more preserved in $\mathcal{M}$ with our approach than without it, i.e. we 
better reconstruct the semantic neighborhood of $\Omega$ in $\mathcal{M}$.
We believe this allows our model to ultimately perform better at cross-modal retrieval.

We finally test the contribution of each component of our proposed loss. We test two variants of our method, where we remove either $\mathcal{L}_{text}$ or $\mathcal{L}_{img}$. We present our results in Table \ref{tab:ablation}. In every case, \textit{combining} our losses for our full method performs the best, suggesting that each loss plays a complementary role in enforcing semantic locality for its target modality.

\subsection{Qualitative results}
In this section, we present qualitative results illustrating how our constraints both improve semantic proximity  and demonstrate superior retrieval results.

\textbf{Semantic proximity: } In Fig.~\ref{fig:relative_distance}, we perform an experiment to discover what samples our constraints affect the most.
We randomly sampled 10k image-image and text-text neighbor pairs (in $\Omega$) and computed their distance in $\mathcal{M}$ using features from our method vs.~the baseline \textsc{Ang+NP+Sym}. Small ratios indicate the samples were closer in $\mathcal{M}$ using our method, relative to the baseline, while larger indicate the opposite. 
We show the samples with the \emph{two smallest} ratios for images and text. We observe that visually dissimilar, but semantically similar images have the smallest ratio (e.g.~E.U.~flag and Merkel, Judge's gavel and Supreme Court), which suggests our $\mathcal{L}_{img}$ constraint has moved the samples closer than the baseline places them. For text, we observe articles about the same  issue are brought closer even though specifics differ.

\textbf{Cross-modal retrieval results: } In Fig.~\ref{fig:cross_modal_retrieval} we show the top-3 results for a set of queries, retrieved by
our method vs.~\textsc{Ang+NP+Sym}. We observe increased semantic homogeneity in the returned samples compared with the baseline. For example, images retrieved for ``drugs'' using our method consistently feature marijuana, while the baseline returns images of pills, smoke, and incorrect retrievals; ``wall'' results in consistent images of the border wall; ``immigration'' features arrests. For text retrieval, we find that our method consistently performs better at recognizing public figures and returning related articles. 
\section{Conclusions}
We proposed a novel loss function which improves semantic coherence for cross-modal retrieval.
Our approach leverages a latent space learned on text alone, in order to enforce proximity within samples of the same modality, in the learned cross-modal space. We constrain text and image embeddings to be close in joint space if they or their partners were close in the unimodal text space. We experimentally demonstrate that our approach significantly improves upon several state-of-the-art loss functions on multiple challenging datasets. We presented qualitative results demonstrating increased semantic homogeneity of retrieval results. Applications of our method include improving retrieval of abstract, non-literal text, visual question answering over news and multimodal media, news curation, and learning general-purpose robust visual-semantic embeddings.

\noindent 
\textbf{Acknowledgements:} 
This material is based upon work supported by the National Science Foundation under Grant No. 1718262. It was also supported by Adobe and Amazon gifts, and an NVIDIA hardware grant. We thank the reviewers and AC for their valuable suggestions.  %

\bibliographystyle{splncs04}
\bibliography{bib}

\end{document}